\documentclass{article}
\usepackage[preprint]{neurips_2025}
\makeatletter
\renewcommand{\@noticestring}{}
\makeatother
\usepackage{float}
\usepackage{graphicx}
\usepackage[utf8]{inputenc}
\usepackage[T1]{fontenc}
\usepackage{hyperref}
\usepackage{url}
\usepackage{booktabs}
\usepackage{amsfonts}
\usepackage{amsmath}
\usepackage{nicefrac}
\usepackage{microtype}
\usepackage{xcolor}
\usepackage[ruled,vlined]{algorithm2e}
\title{On the Quantization Robustness of Diffusion Language Models in Coding Benchmarks\thanks{Code: \url{https://github.com/zinichakraborty/Diffusion-LLM-Quantization-Robustness}}}
\author{%
 Aarav Gupta\thanks{Equal contribution.} \\
 Undergraduate Student \\
 College of Computing\\
 Georgia Institute of Technology\\
 \texttt{aaravgupta@gatech.edu} \\
 \And
 Gururaj Deshpande\footnotemark[2] \\
 Graduate Student \\
 College of Computing\\
 Georgia Institute of Technology\\
 \texttt{gurudesh@gatech.edu} \\
 \AND
 Chandreyi (Zini) Chakraborty\footnotemark[2] \\
 Graduate Student \\
 College of Computing\\
 Georgia Institute of Technology\\
 \texttt{cchakraborty3@gatech.edu} \\
}
\begin{document}

% This has become so pervasive that "vibe coding", referring to the practice of delegating code generation entirely to AI, has emerged as a popular phenomenon.
\maketitle
\begin{abstract}
Auto-regressive Large Language Models (LLMs) achieve strong performance on coding tasks, but incur high memory and inference costs. Diffusion-based language models (d-LLMs) offer bounded inference cost via iterative denoising, but their behavior under post-training quantization (PTQ) has been sparsely explored. We investigate the application and robustness of PTQ techniques, specifically GPTQ and a modified Hessian-Aware Quantization (HAWQ) algorithm, on a diffusion-based coding LLM (CoDA) and observe that these methods applied to CoDA exhibit greater robustness at low bitwidths compared to Qwen3-1.7B, its auto-regressive counterpart, under a standardized evaluation pipeline. We find that in our setup, CoDA exhibits greater robustness at low bitwidths (2-4 bits), with smaller accuracy degradation across HumanEval and MBPP benchmarks. Additionally, mixed-precision configurations derived from HAWQ provide smooth trade-offs across accuracy, latency, and memory. The results suggest that diffusion LLMs may offer advantages for efficient deployment due to more quantization-resilience.
\end{abstract}

\section{Introduction}

Large Language Models, like GPT-4, Gemini, and Claude, are powering new technological innovations \citep{gpt-4, gemini, claude}. Transformer-based autoregressive models (AR models) have shown to be incredibly accurate, display zero-shot performance, and are able to solve challenging benchmarks such as the International Math Olympiad \citep{transformer, foundation_models, few_shot_learners, imo}. However, AR models work by generating one token at a time, which increases memory footprint over long sequences and can also result in long wait times to receive answers.

Several solutions have been proposed and implemented to solve this issue. One of these solutions is Diffusion LLMs (d-LLMs) \citep{sedd, mdlm}, which bring the diffusion training process to text generation. These d-LLMs work by training on the de-noising objective of creating the target sequence over a certain number of steps from a random sequence of tokens. In recent works, d-LLMs have been shown to achieve comparable accuracies to AR models and have tokens-per-second throughput values of over 1000 for simple models \citep{mercury, seed, llada, bd3dm}. However, d-LLMs still suffer the similar issues to AR models of large models requiring large amounts of memory to run as well as long prompts occupying large amounts of memory.

In AR models, one way to overcome these memory issues is by using quantization \citep{first_quant_paper, quantization_survey}. Quantization works by taking the weights of the original model (usually in 32 bit and 16 bit) and examining the scale of the weights of the model to then create lower-bit representations of these weights. During inference, the quantized weights are loaded and de-quantized to the original bit width to complete the forward pass. This is referred to as weight quantization. There exists a large amount of prior work that applies such strategies to LLMs \citep{smoothquant, squeezellm, gptq, awq, spinquant}, but applications of quantization to d-LLMs are quite sparse.

We investigate whether diffusion-based language models are more robust to post-training quantization than autoregressive language models under comparable settings. While we control for model architecture and evaluation pipeline, differences in training data and objectives remain a confounding factor. As a result, differences in robustness may reflect training dynamics rather than generation paradigm alone. We therefore interpret these results as indicative rather than definitive evidence.

Our contributions are the following:
\begin{enumerate}
  % \item We quantize an existing d-LLM using SoTA quantization techniques and examine the pareto-frontier curve of accuracy, latency, and memory usage
  % \item We examine how an existing d-LLM and its AR counterpart place on the pareto-frontier curve showing that d-LLMs are more robust to quantization compared to AR models
  \item We create a standardized empirical comparison of PTQ robustness between a d-LLM and its autoregressive counterpart
  \item We show that the autoregressive model experiences greater performance degradation than its diffusion counterpart under low-bit quantization (2-4 bit) across coding benchmarks
  \item We demonstrate that mixed-precision diffusion LLMs achieve smoother accuracy-latency-memory Pareto tradeoffs in our setting
\end{enumerate}

\section{Related Works}
\subsection{Deep Neural Network Quantization}
Applying quantization to neural networks is a long-standing technique to reduce memory footprint and inference costs by representing weights and activations in a lower numerical precision rather than full 32-bit or 16-bit floating-point formats. Early formulations of neural network quantization formalizing post-training quantization (PTQ) and quantization-aware training (QAT) for integer-only inference were introduced for vision and mobile deployment, where PTQ applies quantization to a fully trained model while QAT incorporates simulated quantization effects during training \citep{jacob_quant,krishnamoorthi_quant,first_quant_paper,quantization_survey}. Building on this framework, mixed-precision methods guided by sensitivity information such as Hessian eigenvalues or traces allocate more bits to layers in the model that are more important for the loss and fewer bits to more robust layers, producing a better Pareto tradeoff between accuracy and compression \citep{hawq,hawq_v2,differentiable_mpq,quantization_survey}. Specifically, Hessian Aware Quantization (HAWQ) uses second-order curvature information to assign per-layer bitwidths \citep{hawq,hawq_v2}. Further work explores broader quantization designs such as uniform vs non-uniform, symmetric vs asymmetric, per-tensor vs per-channel, and weight-only vs weight-plus-activation, often targeting vision models \citep{jacob_quant,krishnamoorthi_quant,quantization_survey}. These base methods can be further adapted to large language models.

\subsection{Large Language Model Quantization}
More recently, quantization has been vastly explored for autoregressive LLMs, where the primary bottleneck comes from memory bandwidth and GPU VRAM rather than FLOPs \citep{zeroquant,llm_int8,gptq,awq,squeezellm,spinquant,spqr}. Early work on LLM quantization include ZeroQuant and LLM.int8(), which showed that 8-bit precision on both weight-only and weight-plus-activation quantization preserved task performance on GPT-style models, allowing single-GPU inference for models that originally required parallelism \citep{zeroquant,llm_int8}. Building off of PTQ, GPTQ introduced a highly effective algorithm tailored to generative transformers, choosing quantized weights such that each layer’s output activations remain as close as possible to the original model, minimizing activation reconstruction errors \citep{gptq}. GPTQ has been applied to a wide range of autoregressive LLMs and can achieve near-lossless 4-bit performance and has become a standard baseline for post-training LLM compression \citep{gptq}. Complementary to this approach, HAWQ-inspired methods extend mixed-precision quantization to LLMs by adapting second-order sensitivity estimates to transformer blocks, enabling per-layer bitwidth allocation without retraining \citep{hawq,hawq_v2,spinquant}. Similar explorations include SqueezeLLM, which combines dense low-precision weights with sparse high-precision outliers, along with broader work on group-wise quantization, block floating-point formats, and hardware-aware bitwidth search to jointly optimize accuracy and end-to-end latency \citep{squeezellm,awq,spqr,spinquant,zeroquant}. Overall, extensive literature demonstrates that LLMs can be aggressively quantized, especially with methods such as GPTQ and HAWQ; however, these techniques have been almost exclusively evaluated on autoregressive LLMs.

\subsection{Diffusion Language Model Quantization}
In contrast, quantization for diffusion-based generative models is much less explored, especially in the language domain. Early work on diffusion quantization focused on vision models, where quantizing U-net backbones in variational autoencoders was applied to speed up image diffusion; for example, Q-Diffusion and related approaches apply uniform or mixed-precision quantization to the score network and show that moderate bitwidth reduction can preserve image quality \citep{qdiffusion,ptq4dm,efficient_diffusion_quant}. These methods are successful on image representations but do not confront challenges specific to text diffusion such as discrete tokenization and bidirectional attention. Very recently, DLLMQuant proposed what is, to our knowledge, the first dedicated PTQ framework for d-LLMs, considering iterative generation, dynamic masking ratios, and the coexistence of stabilized and stochastic tokens make standard PTQ methods fail on d-LLMs \citep{dllmquant}. Specifically, DLLMQuant introduces techniques such as temporally and mask-aware calibration and interaction-aware activation quantization tailored to diffusion-based decoding, showing improved robustness across reasoning and generation benchmarks. However, DLLMQuant largely assumes a uniform precision budget and focuses on stabilizing diffusion-specific behaviors rather than exploring mixed-precision tradeoffs. More broadly, existing diffusion LLM quantization work evaluates diffusion model largely in isolation rather than as a competitor to their autoregressive counterparts.

\section{Experimental Setup}

Our work extends this line of research by applying state-of-the-art post-training quantization methods to both a diffusion LLM (CoDA 1.7B) \citep{coda} and its autoregressive counterpart (Qwen3-1.7B) \citep{qwen3} under a shared, controlled evaluation pipeline. By explicitly characterizing the Pareto frontier over latency, memory footprint, and task performance across multiple standardized coding benchmarks, we provide empirical evidence that diffusion-trained LLMs may exhibit greater robustness than autoregressive models. We will refer to Qwen3-1.7B as Qwen3 throughout this paper.

\subsection{GPTQ}

GPTQ is a weight-only PTQ method that uses a calibration dataset to approximate an inverse Hessian matrix for each layer. In each layer, GPTQ then quantizes weights per-group based on the channel dimension in a fixed order. After quantizing the current set of weights, GPTQ uses the inverse Hessian matrix to update the remaining weights to compensate for the quantization error. During inference, GPTQ follows other weight-only PTQ methods by dequantizing the weights in each layer when needed. This outperforms naive round-to-nearest (RTN) methods as it uses the approximate Hessian to account for quantization errors allowing lower bit (4 bit) GPTQ models to retain most of the accuracy of the original model.

The GPTQ implementation we used was from GPTQModel \citep{gptqmodel} and exposed through Huggingface optimum, an extension of Huggingface transformers \citep{huggingface_transformers}. To work with a version of transformers, we created a fork of GPTQModel and allowed it to build using an older version of transformers. During environment creation, we compile GPTQModel from scratch, which compiles multiple kernels to improve model latency \citep{marlin}.

We use the WikiText dataset \citep{wikitext} as the calibration dataset and quantize to bit widths ${2,3,4,8}$ for both CoDA and Qwen3-1.7B. Group size was set to 128. For each precision, we loaded the base model checkpoint in bfloat16. During GPTQ calibration, we disabled fused attention kernels and used eager attention to ensure stable activation statistics and compatibility with layer-wise quantization hooks. For both models, we had to add hooks in order for the input dimensions provided by GPTQ to work. The hooks were then removed after GPTQ was applied and the model was saved.

\subsection{Hessian Aware Quantization (HAWQ)}
Hessian Aware Quantization (HAWQ) is a quantization approach inspired by a 2019 IEEE paper introducing it as a method for ultra-low quantization of Convolutional Neural Networks like Inception-v3 and ResNet50 \citep{hawq, inception, resnet}. In our research, we found that not only were the intelligent PTQ techniques applied to Diffusion LLMs relatively sparse, but also that this technique especially had not been explored with respect to Diffusion LLMs. It functions via two main algorithms - Sensitivity Calculation and Sensitivity Based Quantization. HAWQ assigns higher precision to layers that are more sensitive to perturbations in the loss surface, as measured by second-order curvature information, and allocates lower precision to more robust layers.

The original algorithm leveraged power iteration to determine the eigenvalues for each block parameter, ranked each block by sensitivity $S_b = \lambda_b / n_b$ where $\lambda_b$ is the Hessian eigenvalues of the block parameter and $n_b$ is the size of the block parameter. After this, bits were assigned to blocks based off of their sensitivity and then each layer block was finetuned, ordered of sensitivity, until convergence. 

We used an adaptation of this algorithm for our own purposes for two main reasons. First off, we wanted to focus the study on post-training specific quantization techniques to minimize the compute overhead of these techniques. Secondly, the size of the models are very different, resulting in an adaptation to the algorithm being necessary. The original paper was written with respect to models like ResNet50 and Inception-v3 with about 25 million parameters each, and we needed to adapt the algorithm to run relatively efficiently with respect to a model about 66x the size (1.7 billion parameter CoDA). To do this, we had to adapt the algorithm to run with a few different approximations to ensure that it could approximate the same algorithm in a more reasonable amount of time. 

First off, we had to implement finite differences to approximate $Hv$, or the Hessian vector product, due to PyTorch \citep{pytorch} autodiff not functioning properly with the Flash Attention \citep{flash} backend of the Diffusion LLM. This increased the time it took to compute the sensitivities of layers, and rippled into the second main adaptation we made to the algorithm - sparse sampling. For each Decoder block in the model, it would've taken hours, if not days, to complete sensitivity calculations with respect to all 1.7 billion model parameters. To address this, we had a hyperparameter $\rho$ defining the sample ratio, or what percentage of the weights we consider for each iteration of sensitivity approximation. Alongside these two modifications to the algorithm, we also modified other hyperparameters like the number of power iterations to reduce the sensitivity generation time. With the algorithm described in [Algorithm \ref{power_iteration}], $\rho=0.1$, and 5 power iterations, we were able to generate sensitivities for CoDA in a little under 2 hours on an NVIDIA H100 by running Algorithms \ref{power_iteration} and \ref{layer_analysis} in sequence over a randomly selected subset of the WikiText dataset \citep{wikitext}.

After computing sensitivities, we assign quantization precision to each module based on what is described in Algorithm \ref{precision_assignment}. In essence, we have a predefined split ratio $(p_{16}, p_8, p_4)$ where $p_{16} + p_8 + p_4 = 1$. We sort modules by sensitivity in descending order, then greedily assign the top $p_{16}$ fraction to 16-bit, the next $p_8$ fraction to 8-bit, and the remaining $p_4$ fraction to 4-bit precision. We use this to target a model memory size and can then optimize the precision makeup to match that while minimizing performance tradeoff. 

We implement per-group symmetric quantization, where weight matrices are partitioned into fixed-size groups (default 128) along the input dimension, and each group is independently scaled and rounded to the target bit-width. This simulated quantization approach preserves floating-point storage while accurately modeling the effects of reduced precision, enabling mixed-precision configurations through layer-specific bit assignments.

Leveraging sensitivities from Algorithms \ref{power_iteration} and \ref{layer_analysis}, and the quantized model that Algorithm \ref{precision_assignment} provides, we ran various experiments on mixed-precision configurations of CoDA described in the Results section to understand how different precisions played into the performance tradeoff outcome. 

\begin{algorithm}[H]
\label{power_iteration}
\caption{Power Iteration for Layer Sensitivity Computation}
\KwIn{Block Parameters $W_i$ for $i = 1, \cdots, m$, perturbation $\epsilon$, sampling ratio $\rho$, iterations $n$}
\For{$i = 1, 2, \ldots, m$}{ \tcp{For each layer block}
  Compute the gradient of $W_i$ by backpropagation, \textit{i.e.}, $g_i = \frac{dL}{dW_i}$\;
  Draw sparse random vector $v$ (same dimension as $W_i$) with sparsity $\rho$\;
  Normalize $v$, $v = \frac{v}{\|v\|}$\;
  \For{$j = 1, 2, \ldots, n$}{ \tcp{Power Iteration}
    $g_+ \gets \nabla_{W_i} L(W_i + \epsilon v)$ \tcp{Perturbed gradient}
    $Hv \gets \frac{g_+ - g_i}{\epsilon}$ \tcp{Hessian-vector product}
    $\lambda_i \gets \|Hv\|$ \tcp{Eigenvalue estimate}
    Normalize and reset $v$, $v = \frac{Hv}{\|Hv\|}$\;
  }
}
\end{algorithm}

\begin{algorithm}[H]
\label{layer_analysis}
\caption{Sensitivity-Based Layer Analysis}
\KwIn{Block-wise Hessian eigenvalues $\lambda_i$ (computed from Algorithm 1), and block parameter size $n_i$ for $i = 1, \cdots, m$.}
\For{$i = 1, 2, \ldots, m$}{ \tcp{Compute Sensitivity}
  $S_i = \lambda_i$ \tcp{Sort by raw eigenvalues to understand sensitivity} 
}
Order $S_i$ in descending order to determine relative sensitivity ranking for each block.\;
\end{algorithm}

\begin{algorithm}[H]
\label{precision_assignment}
\caption{Sensitivity-Based Precision Assignment}
\KwIn{Module sensitivities $s_m$ for $m = 1, \cdots, M$, split ratios $(p_{16}, p_8, p_4)$ where $p_{16} + p_8 + p_4 = 1$.}
Sort modules by $s_m$ in descending order\;
$k_{16} \gets \lfloor p_{16} \cdot M \rfloor$ \tcp{Cutoff indices}
$k_8 \gets \lfloor (p_{16} + p_8) \cdot M \rfloor$\;
\For{$m = 1, 2, \ldots, M$}{
  \uIf{$m \leq k_{16}$}{$\text{precision}_m \gets 16$}
  \uElseIf{$m \leq k_8$}{$\text{precision}_m \gets 8$}
  \Else{$\text{precision}_m \gets 4$}
}
\end{algorithm}

\subsection{Evaluations}

To evaluate the baseline models (no quantization) and quantized models, we followed a similar evaluation recipe from CoDA \citep{coda} where they evaluated models on HumanEval \citep{humaneval} and Mostly Basic Programming Problems (MBPP) \citep{mbpp}. Both datasets report a pass@1 and pass@5 value where pass@1 refers to the model passing all test cases in one generation, and pass@5 refers to the model passing all test cases if one generation passes all test cases from five generations. We evaluate on the regular dataset as well as the plus dataset using the Language Model Evaluation Harness and specifically use pass@1 as our evaluation metric \citep{eval-harness}.

To measure per-step model latency, we execute each model on a single text input of sequence length 1024 on an NVIDIA L40S GPU. For CoDA, we measure the latency of a single denoising step, and for Qwen3, we measure the latency of generating a single token. We perform 200 warm-up runs, followed by 2000 timed runs to compute the mean latency and standard deviation.

\section{Results and Discussion}

\subsection{Accuracy of CoDA vs Qwen3}

Baseline and GTPQ quantized benchmark evaluation results are shown in Table \ref{tab:coda_qwen_results}.

\subsubsection{Baseline}

We observe that Qwen3's base model performed better across latency and accuracy in comparison to our CoDA model. This observation is also supported by the results in the CoDA paper \citep{coda}. The main reason this could potentially happen is due to dataset and training differences. Qwen3 does not publicly release their pre-training or post-training dataset and does not release their training hyperparameters. However, the Qwen3 technical report states that pre-training was done on 36 trillion tokens whereas CoDA was only pre-trained on 180 billion tokens. It's possible the inclusion of more data allowed Qwen3 to contain stronger representations than CoDA.

\subsubsection{GPTQ Quantized}

We observe that CoDA is able to retain benchmark accuracy all the way down to 4 bit precision before it starts to collapse at 2 and 3 bit precision. Qwen3 also shows a similar trend of retaining benchmark accuracy down to 4 bit precision but complete model collapse happens at 3 bit precision. We see that CoDA at 4 bit precision only loses 1.8 points of accuracy at best (observed on HumanEval Plus) and 5 points of accuracy at worst (observed on MBPP Instruct) while Qwen3 at 4 bit precision loses 12 points of accuracy at best (observed on MBPP Instruct) and loses 23.2 points of accuracy at worst (observed on HumanEval Instruct). This is also visible in Figure \ref{pareto_frontier}. This suggests that CoDA retains accuracy better than Qwen3 under these settings, providing evidence that diffusion-based LLMs may be more robust to quantization in this experimental setup.

A possible explanation, and a counterargument, to d-LLMs being robust to quantization, is supported by recent work focusing on the scaling laws of transformer-based language models at precision \citep{scaling_laws_for_precision}. This work observed that PTQ techniques can degrade over-trained models, which they define as models that were trained with a high data to parameter ratio, more than models that follow the Chinchilla-optimal \citep{chinchilla} data to parameter ratio. Since for both pre-training and post-training, Qwen3 was trained on more data than CoDA, and CoDA and Qwen3 have the same number of parameters, then this means that Qwen3 was trained with a higher data to parameter ratio worsening its performance under PTQ techniques. However, this work only trains models up to 300 million parameters using an OLMo-style, and it is unclear whether this exists for larger models with different architectures.

\subsection{Latency of CoDA vs Qwen3}

Baseline and GTPQ quantized benchmark evaluation results are shown in Figure \ref{fig:latency_figure}.

\subsubsection{Baseline}

For the baseline model, the mean latency for Qwen3 was 26.843 ms with a standard deviation of 0.305 while for CoDA the mean latency was 28.329 with a standard deviation of 0.305.

We observe that the baseline Qwen3 model has a slightly lower mean latency than CoDA. This is likely due to single-token AR decoding being highly optimized through causal masking as well as benefiting greatly from highly-optimized kernels. In contrast, each denoising step in CoDA operates over the full sequence with bidirectional attention, resulting in dense per-step computation. These architectural and kernel-level differences result in a modest but consistent latency advantage for Qwen3 in the 16-bit baseline setting.

\subsubsection{GPTQ Quantized}

% We found that while the base CoDA model performed worse than Qwen3, the GPTQ quantized models all had generally lower latency.

We observe that the latency for the quantized CoDA and Qwen3 models are very similar to each other. We noticed that the latency of CoDA with GPTQ quantization at 4 and 8 bit is similar to the latency of CoDA at 16 bit. For Qwen3, we see that at 4 and 8 bit, the latency is a bit higher than the baseline model by around 5-10 milliseconds. The reason that 4 and 8 bit model latency is very close to the 16 bit baseline is due to the Marlin kernel being used \citep{marlin}. The Marlin Kernel only exists for 4 and 8 bit precision, so we see that at 2 and 3 bit, the latency is higher as the Torch Quant Linear or the Triton Quant Linear kernels are used, which are slower than the Marlin kernel \citep{pytorch, triton, marlin}.

After applying GPTQ, CoDA becomes competitive with Qwen3 and surpasses it at 2bit, 4bit, and 8bit precision, despite being slightly slower at its baseline. This behavior is likely due to systems-level overheads that become more prominent once arithmetic cost is reduced due to quantization. 

\begin{table}[H]
  \centering
  \caption{Performance of different quantization levels on HumanEval and MBPP.  
  Each entry shows the CoDA result, with the Qwen3 result in parentheses.}
  \label{tab:coda_qwen_results}
  \begin{tabular}{lcccc}
    \toprule
    Quantization & \multicolumn{2}{c}{HumanEval} & \multicolumn{2}{c}{MBPP} \\
    \cmidrule(r){2-3} \cmidrule(r){4-5}
    & Instruct & Plus & Instruct & Plus \\
    \midrule
    2-bit             & 0.000 (0.000) & 0.000 (0.000) & 0.000 (0.000) & 0.000 (0.000) \\
    3-bit             & 0.317 (0.000) & 0.292 (0.000) & 0.362 (0.002) & 0.479 (0.000) \\
    4-bit             & 0.457 (0.439) & 0.421 (0.409) & 0.418 (0.358) & 0.574 (0.503) \\
    8-bit             & 0.481 (0.665) & 0.421 (0.616) & 0.466 (0.490) & 0.632 (0.656) \\
    16-bit (baseline) & 0.481 (0.671) & 0.439 (0.628) & 0.468 (0.478) & 0.619 (0.664) \\
    \bottomrule
  \end{tabular}
\end{table}

We see more significant drops in accuracy with increased quantization between Qwen3 and CoDA (Table \ref{tab:coda_qwen_results}). Specifically, at 3-bit quantization there is a catastrophic drop in accuracy across the benchmarks for the autoregressive model, while the diffusion LLM maintains some functionality. We also see more robustness at 4-bit quantization for diffusion LLMs with a smaller drop in accuracy.

\begin{figure}[H]
  \centering
  \includegraphics[width=0.7\linewidth]{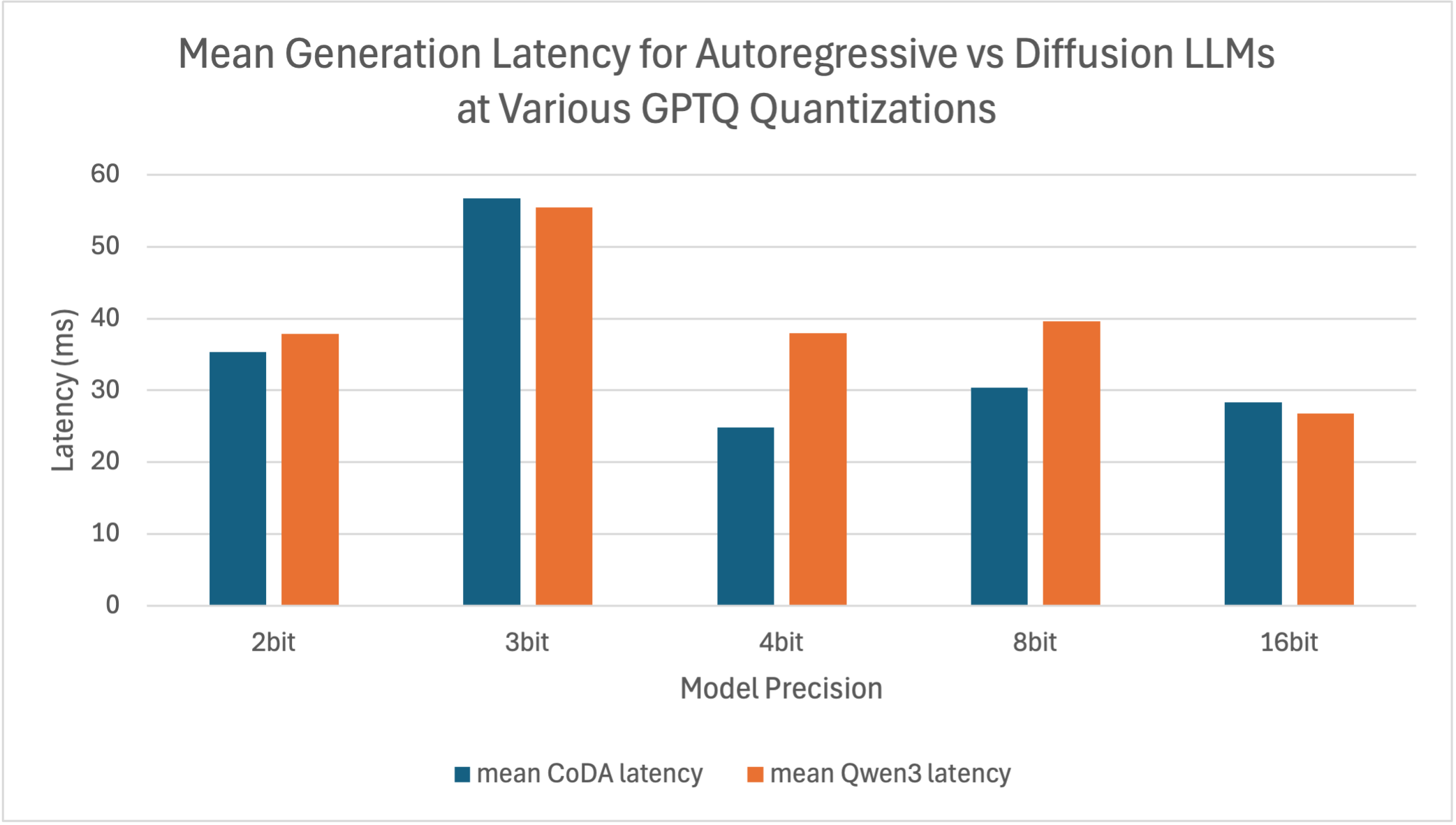}
  \caption{Graphic of Latency-Precision Tradeoff between CoDA and Qwen3 GPTQ Models}
  \label{fig:latency_figure}
\end{figure}

Figure \ref{fig:latency_figure} shows that while baseline latency favors Qwen3, quantized CoDA models match or outperform Qwen3 at lower bitwidths, indicating that system-level overheads dominate as arithmetic cost decreases.

While CoDA executes a single denoising forward pass, Qwen3's autoregressive path involves framework-level overhead such as causal attention setup, token position handling, and output projection logic. In contrast, CoDA's denoising forward pass is a single, self-contained computation over the sequence. As quantization reduces the cost of underlying matrix operations, these fixed overheads account more to the total latency, allowing CoDA to exceed Qwen3's performance at lower bitwidths.

\subsection{Hessian Aware Quantization (HAWQ)}
Building off of the results from GPTQ, we ran several experiments using our HAWQ implementation to explore the Pareto Frontier of the performance-memory footprint tradeoff at a more granular level. We started with a few fairly basic configurations, specifically 16/8 bit (50\% most sensitive are 16bit, rest are 8bit) and 8/4 bit (same as 16/8 but 8 bit and 4 bit). In the below figure, we found that HAWQ treads the frontier very well, and allows users to target a much more specific memory footprint target for different types of devices. Looking at Figure 1, we can see that the model has minimal performance tradeoff on the benchmark evaluations between different precisions when split 50/50 in precision, with the higher precisions assigned to the more sensitive weights. 

\begin{figure}[H]
  \centering
  \includegraphics[width=1\linewidth]{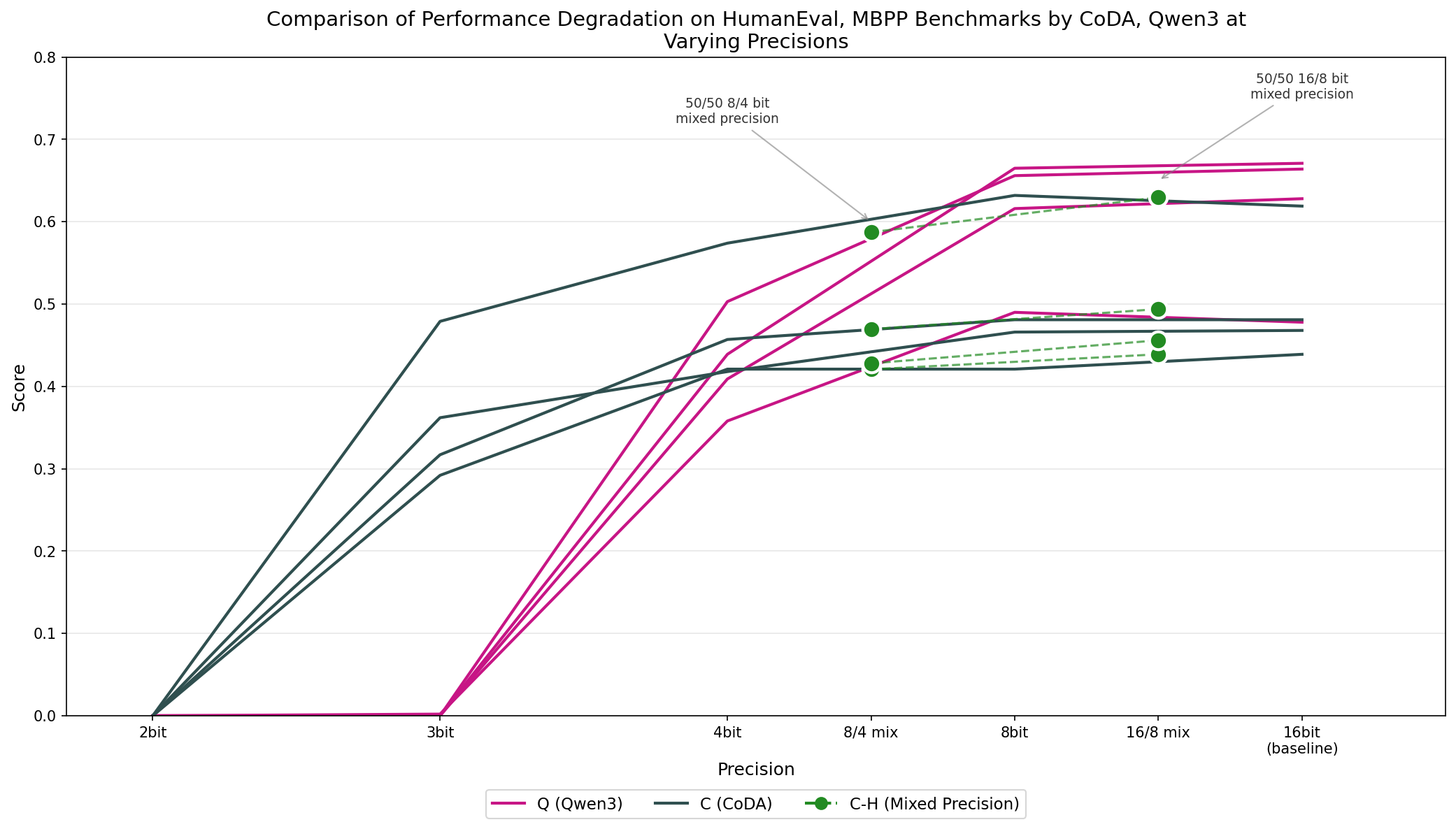}
  \caption{Graphic of Performance-Precision Tradeoff with HAWQ Performance Labeled}
  \label{pareto_frontier}
\end{figure}

Figure \ref{pareto_frontier} illustrates that mixed-precision configurations closely track the Pareto frontier, enabling fine-grained tradeoffs between accuracy and memory usage with minimal degradation at moderate compression levels.

However, it appears that this initial approach suffers with more complex mixes of precision. For example, we tried several configurations combining 16bit, 8bit, and 4 bit precisions based on model block sensitivities and continually saw the model collapse on all 4 benchmarks when evaluated, likely due to the model being unable to correct itself due to imprecise weights and no method for the model to recover performance. It is possible that additional post-training or supervised fine-tuning on the model, suggested in the original Hessian-Aware Quantization paper, could assist with performance recovery post quantization. We elaborate on what we believe are the best ways to continue our work in exploring the applications and benefits of PTQ techniques for Diffusion LLMs in Section \ref{future_work}.

\section{Future Work}
\label{future_work}
Looking forward, a natural next step would be to train a diffusion and autoregressive model from scratch on the same data, keeping the same tokenizer, architecture, and hyperparameters to ensure even more equal footing when comparing the models to further validate our results. Because the training data for Qwen3 was not released, it's very challenging to analyze the data it was trained on and how that may ripple out to the robustness of the model. Our experiments show that it's likely a factor related to model architecture, but training two models on the exact same datasets with everything except for generation mode (diffusion vs autoregressive) identical could help truly isolate the difference in performance, robustness, and quality that the generation paradigm provides. 

Secondly, we would want to explore the impact of calibration datasets on the model performance. Both models were calibrated on WikiText \citep{wikitext}, which we believed would expose the models to a wide variety of content and judge the sensitivity of model layers to the general knowledge base. In a future iteration of this project, we aim to use a subset of a dataset like OpenCoder \citep{opencoder}, with an explicit coding focus to understand how that impacts the effects of quantization and measured layer sensitivities. 

Thirdly, we'd like to explore HAWQ's execution more thoroughly. Due to time constraints, we were unable to create and evaluate as many permutations of the CoDA model with mixed precision in accordance with HAWQ's algorithm as we would have liked to, as the process of tracking sensitivities, generating mixed precision models, and then sending off evaluation jobs is very slow, time consuming exercise. If this project was continued, we'd like to further explore how different combinations of precisions affect performance, and what patterns may lead to more consistent performance preservation even at smaller model sizes targeting tighter memory footprints. We found that splits between two precisions worked very well at treading the Pareto Frontier, but would like to explore more thoroughly how mixes between 3 or more precisions perform, alongside the best ways to minimize lost performance.

Lastly, we'd also like to explore the impact of quantization-aware training (QAT). PTQ has shown strong potential in our experiments to reduce model size while still retaining strong model performance, but it's possible that some kind of training could result in performance recovery. For example, it's possible that the model collapse we observed at 4-bit and lower precision could be alleviated with training for performance recovery, as observed in the original HAWQ paper for example. 

\section{Conclusion}
We compare the effects and tradeoffs of PTQ on Qwen3-1.7B and Salesforce's CoDA 1.7B, trained in a diffusion fashion on Qwen3's architecture. We find that with GPTQ used to quantize both models across varying precisions, CoDA's mean generation latency is on par with that of Qwen3 across the HumanEval and MBPP evaluation datasets, and meaningfully faster by 25\%-40\% at 4 and 8 bit precisions with GPTQ used as the optimization method. We also find that CoDA is much more robust in terms of the performance-precision tradeoff. While Qwen3's performance collapsed across all 4 coding benchmarks at precisions of 4-bit and lower, with a an average drop of 40\% per benchmark in performance when quantized from 16-bit to 4-bit, while CoDA only experienced an average drop of about 8\% per benchmark with the same change in precision. Our experiments with HAWQ show that it smoothly forms a Pareto Frontier for the performance-precision tradeoff. The results suggest PTQ techniques on diffusion-based language models have lower performance degradation at low-bit precisions due to CoDA's robustness to collapse at lower precisions compared to Qwen3 and offer a promising direction for efficient and robust language model deployment under resource constraints.

\section{Acknowledgements}

This research was supported in part through research cyberinfrastructure resources and services provided by the Partnership for an Advanced Computing Environment (PACE) \citep{PACE} at the Georgia Institute of Technology, Atlanta, Georgia, USA. Our work wouldn't be possible without the GPUs provided by them. 

We'd also like to thank Professor Tumanov, alongside Dhruv Garg and Sukrit Kumar, for their guidance and feedback throughout the course of this project in CS 8803: Systems for Machine Learning. 

\bibliographystyle{plainnat}
\bibliography{references}
\end{document}